\documentclass[a4paper, 10pt, conference]{ieeeconf}      
\usepackage{FG2025}

\usepackage{hyperref}
\usepackage{graphicx}
\usepackage{booktabs}
\usepackage{multirow}   

\FGfinalcopy 

\IEEEoverridecommandlockouts                              
\def\FGPaperID{320} 

\title{\LARGE \bf
Comparison of Visual Trackers for Biomechanical Analysis of Running
}

\author{\parbox{16cm}{\centering
    {\large Luis F. Gomez$^1$, Gonzalo Garrido-Lopez$^2$, Julian Fierrez$^1$, Aythami Morales$^1$, \\Ruben Tolosana$^1$, Javier Rueda$^2$ and Enrique Navarro$^2$}\\
    \vspace{3mm}
    {\normalsize
    $^1$ BiDA Lab, Universidad Autónoma de Madrid, Madrid, Spain\\
    $^2$ Sport Biomechanics Laboratory, Faculty of Physical Activity and Sports Sciences
INEF, Universidad Politécnica de Madrid, Madrid, Spain}}
    \thanks{Funding from BBforTAI (PID2021-127641OB-I00 MICINN/FEDER). Morales is supported by the Madrid Government under the Multiannual Agreement with UAM in the line of Excellence for the University Teaching Staff (V PRICIT). Garrido is supported by Banco Santander under a doctoral grant. Thanks to Alejandro Carrillo for his work processing the data.}
}

\usepackage{fancyhdr}
\begin{document}

\ifFGfinal
\thispagestyle{empty}
\pagestyle{empty}
\else
\author{Anonymous FG2025 submission\\ Paper ID \FGPaperID \\}
\pagestyle{plain}
\fi
\maketitle

\thispagestyle{fancy} 

\begin{abstract}
Human pose estimation has witnessed significant advancements in recent years, mainly due to the integration of deep learning models, the availability of a vast amount of data, and large computational resources. These developments have led to highly accurate body tracking systems, which have direct applications in sports analysis and performance evaluation.
This work analyzes the performance of six trackers: two point trackers and four joint trackers for biomechanical analysis in sprints. The proposed framework compares the results obtained from these pose trackers with the manual annotations of biomechanical experts for more than 5870 frames. The experimental framework employs forty sprints from five professional runners, focusing on three key angles in sprint biomechanics: trunk inclination, hip flex extension, and knee flex extension. We propose a post-processing module for outlier detection and fusion prediction in the joint angles.
The experimental results demonstrate that using joint-based models yields root mean squared errors ranging from 11.41° to 4.37°. When integrated with the post-processing modules, these errors can be reduced to 6.99° and 3.88°, respectively. The experimental findings suggest that human pose tracking approaches can be valuable resources for the biomechanical analysis of running. However, there is still room for improvement in applications where high accuracy is required.
\end{abstract}

\section{Introduction}
\label{sec:Introduction}

\setcounter{footnote}{0} 
The rapid evolution of computer vision in recent years has resulted in numerous opportunities in many fields. Advancements in human tracking technologies employ machine learning, deep learning, and sensor fusion to improve motion analysis's precision, scalability, and efficiency. These developments shape sports science by allowing precise performance assessments, injury prevention strategies, and individualized training programs. Modern tracking systems help athletes and coaches optimize techniques and enhance overall athletic performance by providing real-time information.

Sprinting is a fundamental aspect in many sports and is crucial in determining athletic success~\cite{morin2011technical}. Understanding sprint mechanics is essential for both performance enhancement and injury prevention. 
Historically, movement assessment relied on visual observation by experts~\cite{yang2024improving}. However, such methods are subjective and depend on the expertise of the evaluator~\cite{hii2023automated,yang2024improving}. 
To overcome these limitations, manual annotation techniques were introduced to provide objective data, allowing measurement of temporal variables and joint kinematics analysis~\cite{bissas2022kinematic,hanley2022biomechanics}. But these systems are time-consuming and susceptible to human error during video analysis~\cite{cronin2024feasibility}.

Technology integration has led to the development of automated tools for motion analysis. Inertial Measurement Units (IMU) and marker-based motion capture systems have been widely used, the latter offering high precision but requiring controlled environments, trained personnel, and expensive equipment~\cite{paula2023gait,viswakumar2022development,yang2024improving}. However, IMUs face challenges such as sensor drift, and their precision in joint angle estimation remains questionable~\cite{acien2020sensors,bastiaansen2020inertial,lin2023validity,nazarahari2022foot}.

To provide a more practical and cost-effective alternative, smartphone-based applications have emerged as a tool for sprint analysis. However, these applications do not offer detailed joint kinematic data, creating a gap in the precise biomechanical evaluation~\cite{romero2017sprint}.
Markerless motion capture has become an increasingly popular alternative. Using advanced deep learning-based pose estimation techniques, it can track movement from video recordings. 
These systems balance cost and efficiency, making them attractive for sports analysis. However, challenges remain, particularly with regard to high computational demands, the need for manual supervision, and biomechanical precision~\cite{stenum2021two,yang2024improving}.

Our work focuses on the study of sprint mechanics using markerless motion capture systems and how video-based pose estimation can enhance biomechanical analysis. The main aim is to compare currently available state-of-the-art visual trackers of different nature to investigate if they can achieve (perhaps with some modifications) adequate precision of joint kinematics for practical biomechanical analyses (e.g. for injury prevention). For that purpose, we will use as a baseline the VideoRun2D system and dataset\footnote{\url{https://videorun2d.com/}}.

\subsection{Related Works}

Different studies have explored the feasibility of markerless systems in sprint biomechanics, with early solutions such as Microsoft Kinect showing potential but struggling with accurate joint angle estimation in low light conditions and the clothes of the subjects~\cite {ota2021verification,takeichi2018mobile,viswakumar2022development}.

Today, advanced human pose estimation models rely on deep neural networks to detect and track body landmarks from video footage~\cite{acien2020sensors,viswakumar2022development,yang2024improving}. Although these systems perform well in visual tracking, their biomechanical accuracy remains a concern. 
In 2022, the research conducted by Chung \textit{et al.}~\cite{chung2022comparative} and Jo and Kim ~\cite{jo2022comparative} compared the efficiency of four human pose estimators in various sports scenarios. However, the study focused solely on the prediction precision of the joint points without a biomechanical analysis. The results indicated that pose estimation models can achieve an accuracy between 70\% and 80\%. In 2024, Yan and Park~\cite{yang2024improving} analyzed the angles of the hip, knee, and ankle in a multicamera recording setup. The study compared marker and markerless systems and reported values below the mean absolute error of 5°. In the same year, Galasso \textit{et al.} in~\cite{galasso2024novel} compared IMU systems and OpenPose~\cite{cao2019openpose}, a human pose estimator. They analyzed 21 subjects and evaluated the precision of both systems. The results achieved a precision of 93\%, confirming that markerless approaches can complement or even substitute traditional IMU systems in biomechanical evaluations in some cases.

These studies definitively show the potential of markerless estimators for biomechanical analysis. However, challenges remain. For example, many deep learning models have not yet been employed in biomechanical running analysis. In addition, the precision required for medical purposes, such as preventing runners from injury, is not yet sufficient, as an accuracy close to 1.9° is needed~\cite{lee2009running}. 

\subsection{Contribution of this Work}

As shown in the review of the literature, the use of deep learning-based pose estimation models in biomechanical analysis remains an active area of research. To contribute to this field, we investigate multiple state-of-the-art pose estimation approaches with improved signal post-processing compared to expert manual annotations. The main contributions of this work are as follows.

\begin{itemize}
    \item We analyze six approaches to estimation of human pose based on deep learning models for sprint biomechanics analysis.
    \item We evaluated each markerless system against a manually marked ground truth to quantify biomechanical errors across different configurations.
    \item We propose a pipeline to correct common visual tracking errors encountered in challenging recording scenarios, specifically when analyzing runners from lateral views.
\end{itemize}

These contributions aim to improve cost-effective sprint analysis and can benefit various fields, including professional sports monitoring, injury rehabilitation, and forensic applications. This research has the potential to drive advances beyond sports science, enhancing motion capture accuracy and biomechanical validity.

\section{Materials and Methods}
\label{sec:ProposedSystem}

\subsection{Dataset}

This study uses the publicly available VideoRun2D dataset, originally presented by Garrido \textit{et al.} in~\cite{garrido2024videorun2d}. The dataset comprises video recordings of short-distance sprints performed by five healthy amateur soccer players. Each participant completed eight 20-meter sprints on natural grass fields, with videos capturing the 5th to 15th meter segment using a high-speed camera (1920$\times$1080 pixels, 100 fps). The camera was placed perpendicular to the sagittal plane to ensure a side view of the motion, following the recording protocol outlined by Yang and Park~\cite{yang2024improving}.
In total, the dataset contains 40 sprint trials, 513 annotated running strides, and 5,870 video frames~\cite{garrido2024videorun2d}.

\begin{figure*}[t!]
    \centering
    \includegraphics[width=0.6\textwidth]{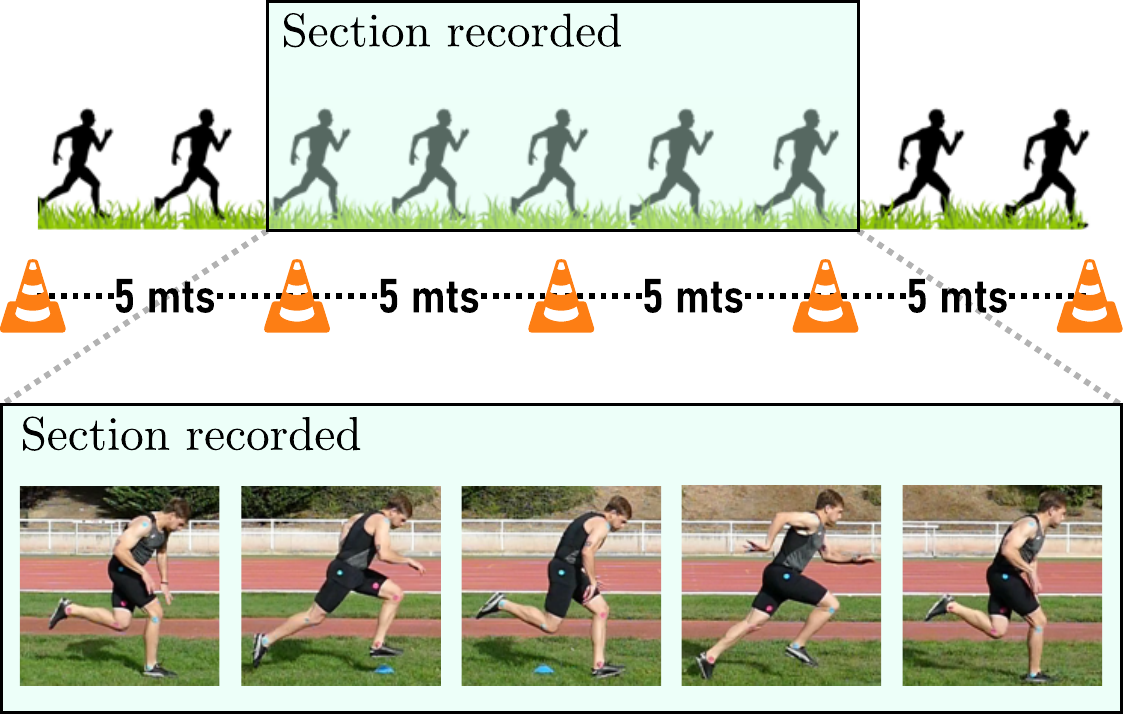}
    \caption{Data acquisition sketch and example of the different stages of a stride for an individual in the dataset.}
    \label{fig:Recording}
\end{figure*}

\subsubsection{Ground truth based on manual labeling}

\begin{figure*}[t!]
    \centering
    \includegraphics[width=0.5\textwidth]{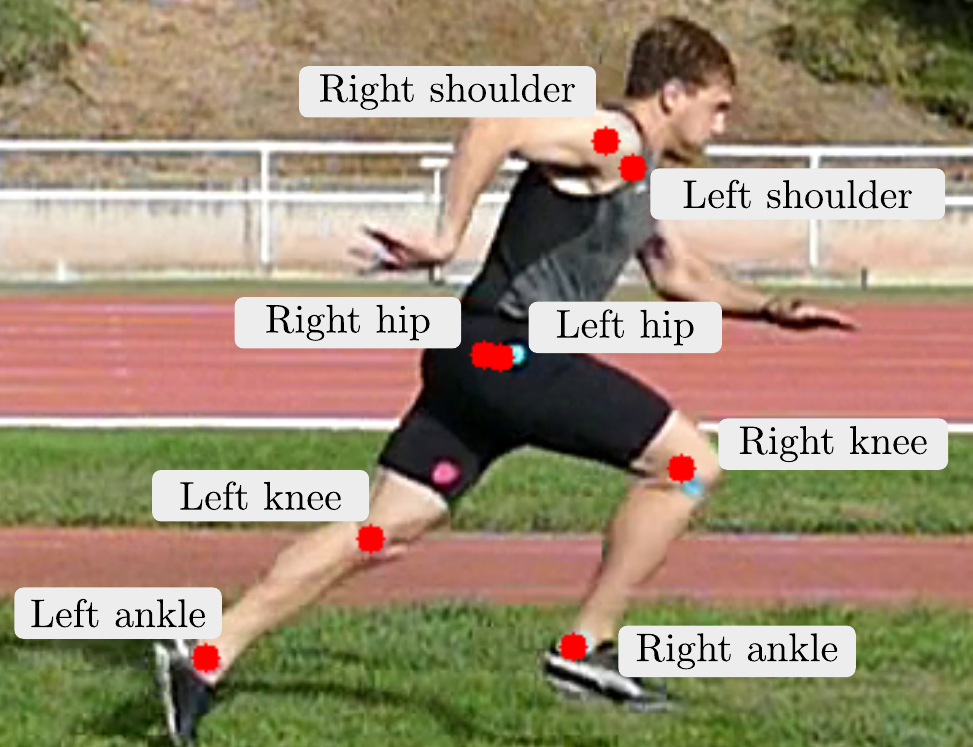}
    \caption{Location of the joint points marked on Kinovea software (ground truth).}
    \label{fig:JointPoints}
\end{figure*}

The VideoRun2D dataset includes manual annotations of joint positions provided by three trained biomechanical experts using the Kinovea tool (version 0.9.5)\footnote{ \href{https://www.kinovea.org/}{https://www.kinovea.org/}}.
The key joints (shoulder, hip, knee, and ankle) were labeled on every frame. When joints were occluded, the positions were estimated based on anatomical symmetry. The annotated trajectories were then reviewed for consistency~\cite{garrido2024videorun2d}. An illustration of the location of the labeled joints is shown in Figure~\ref{fig:JointPoints}.

\subsection{VideoRun2D: System}

\begin{figure*}[t!]
    \centering
    \includegraphics[width=\textwidth]{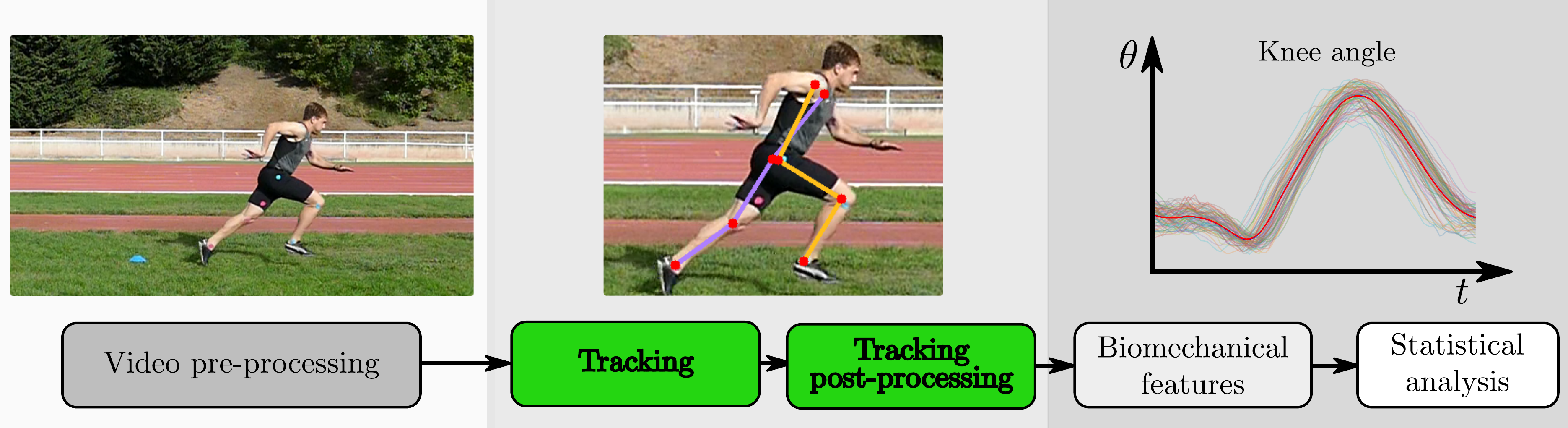}
    \caption{Block diagram of the VideoRun2D system. The system has five modules that estimate joint angles. }
    \label{fig:BlockDiagram}
\end{figure*}

VideoRun2D performs markerless body tracking and estimates the joint angles of each user during a sprint~\cite{garrido2024videorun2d}. The system comprises five processing modules: video pre-processing, tracking of the joint points, tracking post-processing, generation of biomechanical features, and a validation system that employs statistical analysis. 
Figure~\ref{fig:BlockDiagram} shows a diagram of the VideoRun2D system.

\subsection{VideoRun2D: Modules}

\subsubsection{Video pre-processing}

This module depends on the quality~\cite{q12fer} and characteristics of the input video and the behavior of the following tracking module. In this study, the input video is of very high quality~\cite{q22survey}, and the following trackers are highly robust to typical input covariates (e.g., illumination changes, typical video noise such as coding artifacts, etc.). 
Therefore, our pre-processing pipeline comprises only person detection without specific landmarks (square crop), reducing the input image from $1920\times1080$ to $300\times300$ (see center picture in Figure~\ref{fig:BlockDiagram}).
More elaborate pre-processing can be considered for VideoRun2D applications using another kind of less-quality image input~\cite{super19fer} or other trackers in the following module more sensible to particular image covariates~\cite{q12fer}.

\subsubsection{Tracking} \label{sec:Tracking}

As the core of the body tracker in VideoRun2D, we have considered six competitive trackers:
CoTracker2~\cite{karaev2023cotracker}, CoTracker3~\cite{karaev2024cotracker3}, PoseNet~\cite{papandreou2018personlab}, MediaPipe~\cite{ohri2021device}, MoveNet~\cite{movenet2021} and MMPose~\cite{mmpose2020}. 

\begin{itemize}

\item CoTracker2 and CoTracker3 are point tracking models proposed by Karaev~\textit{et al.} in~\cite{karaev2023cotracker} and~\cite{karaev2024cotracker3}. They are based on transformers~\cite{paula2023gait} and attention neural networks and track a dense grid of points in a video sequence. Although the CoTracker system is semi-supervised, only the initial and final points of the run need to be provided to determine the trajectory in the video. Both CoTracker versions were implemented with the default parameters on their project page\footnote{\href{https://cotracker3.github.io/}{https://cotracker3.github.io/}}.

\end{itemize}

The other 4 trackers are all based on automatic joint point tracking, which locates the shoulder, elbow, hand, hip, knee, and ankle joint centers on the right and left sides of the human body.

\begin{itemize}

\item PoseNet is a model that Google developed in 2018~\cite{papandreou2018personlab}. It is a model that allows for the estimation of 17 landmarks around the human body. The architecture is based on convolutional neural networks optimized for mobile devices and browsers, focusing on lightweight applications such as physical activity monitoring and interactive entertainment.

\item MediaPipe, developed by Google in 2019~\cite{ohri2021device, xu2020ghum}, is a framework that integrates deep learning methodologies and provides various solutions for applications such as face, iris, hand, and pose detection. The pose recognition module detects 33 landmarks, classified into 12 on the body, 11 on the face, and 10 on specific details of the hands and feet. A distinguishing feature of MediaPipe is its lightweight architecture, a design choice that facilitates its deployment, including on mobile devices, for diverse applications from sports analysis to interactive entertainment. This architecture is analogous to PoseNet, which prioritizes lightweight operation to ensure efficient execution on various platforms.

\item MMPose, introduced in 2020 as part of OpenMMLab~\cite{mmpose2020}, is also a PyTorch-based framework used for estimating human pose. It supports multiple architectures and tasks, including 2D and 3D pose detection in single-person and multi-person estimation.
Its modular and extensible design allows model customization and training for specific applications. At the same time, its high precision makes it suitable for complex projects such as biomechanics, 3D reconstruction, and sports analysis.
In this study, the Real-Time Multi-Person Pose Estimation (RTMPose) method proposed in 2023 by Jiang \textit{et al.} in~\cite{jiang2023rtmpose} is implemented using the MMPose framework.

\item Finally, MoveNet, developed by Google in 2021, is a human pose estimation tool based on a bottom-up estimation model. It uses heat maps to accurately locate human body keypoints. Google and IncludeHealth are using it in TensorFlow Lite for modern desktops, laptops, and phones. 
MoveNet identifies 17 key body points with high precision. It is available in two variants: Lightning, designed for high performance on mobile and edge devices, and Thunder, which provides enhanced accuracy for more demanding use cases.
Due to its high landmark precision, we implemented the MoveNet-Thunder model in this work\footnote{\href{https://www.kaggle.com/models/google/movenet}{https://www.kaggle.com/models/google/movenet}}.

\end{itemize}

\begin{figure*}[!t]
    \centering
    \includegraphics[width=0.95\textwidth]{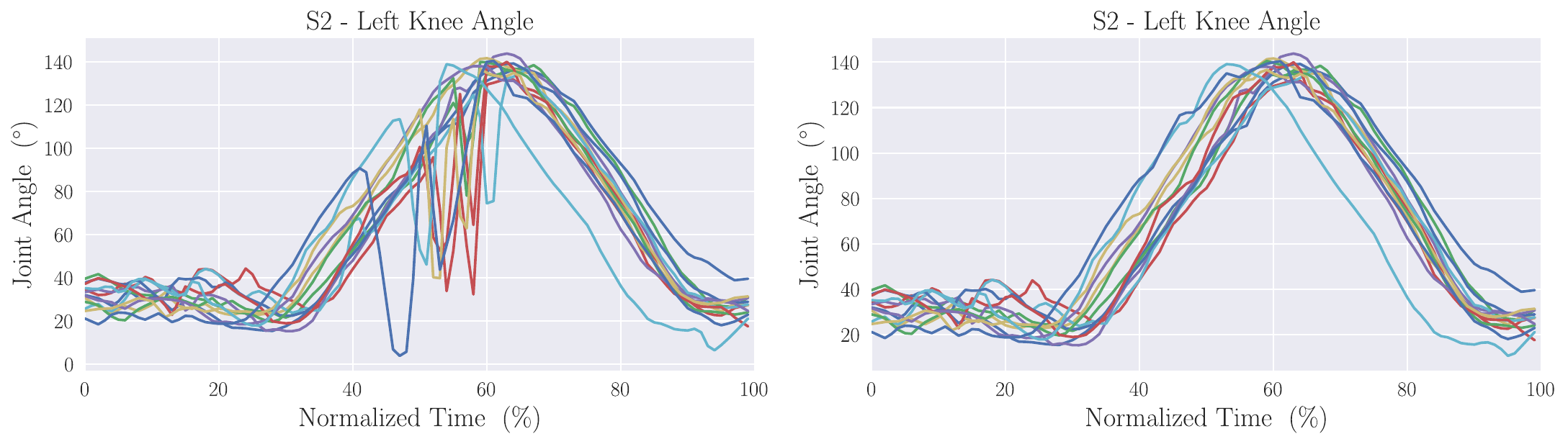}
    \caption{Graphical example of the correction of the left knee angle in all the thirteen strides from the sprinter two, both before (on the left) and after (on the right) post-processing}
    \label{fig:PP_UserStride}
\end{figure*}

\subsubsection{Tracking post-processing}\label{sec:PP_TrackingModule}

After running the evaluated trackers, three types of errors were found in the knee and ankle joints due to occlusions and other confusions in the sagittal plane: 1) loss of points, 2) confusion between the left and right sides of the body, and 3) misallocation of a body point. 

To correct the errors associated with the confusion between the left and right sides, we first smooth the output trajectories of the different trackers $x_{j}(t)$ and $y_{j}(t)$ using support vector regression (SVR) to generate the smoothed trajectories $\widetilde{x}_{j}(t)$ and $\widetilde{y}_{j}(t)$, where $j$ represents the body part tracked and $t$ the time. The principal advantage of using an SVR is the $\epsilon$ margin that makes the regression as robust as necessary to outliers for a particular setup.

Next, we calculate the error curve $e_{m,j} = \left\|  m_{j}(t) - \widetilde{m}_{j}(t)\right\|_{2}$, where $m_{j}(t)$ is either $x_{j}(t)$ or $y_{j}(t)$. We then determine the outlier points at times $t_{\textrm{out}}$ that are above $3\times\sigma_{e}$ where $\sigma_{e}$ is the standard deviation of the error distribution.
Then, the outlier points at those $t_{\textrm{out}}$ are corrected.

Initially, for outliers, we first interchange between the left and right sides. If the outlier is not corrected this way, then the SVR-smoothed data are considered instead of the tracker result, i.e.: $m_j(t_{\textrm{out}}) = \widetilde{m}_j(t_{\textrm{out}})$. 
Figure~\ref{fig:PP_UserStride} shows a tracking result before and after the proposed post-processing for the angle of the left knee of the 13 strides of a given sprinter. 

\begin{figure*}[htpb]
    \centering
    \includegraphics[width=0.92\textwidth]{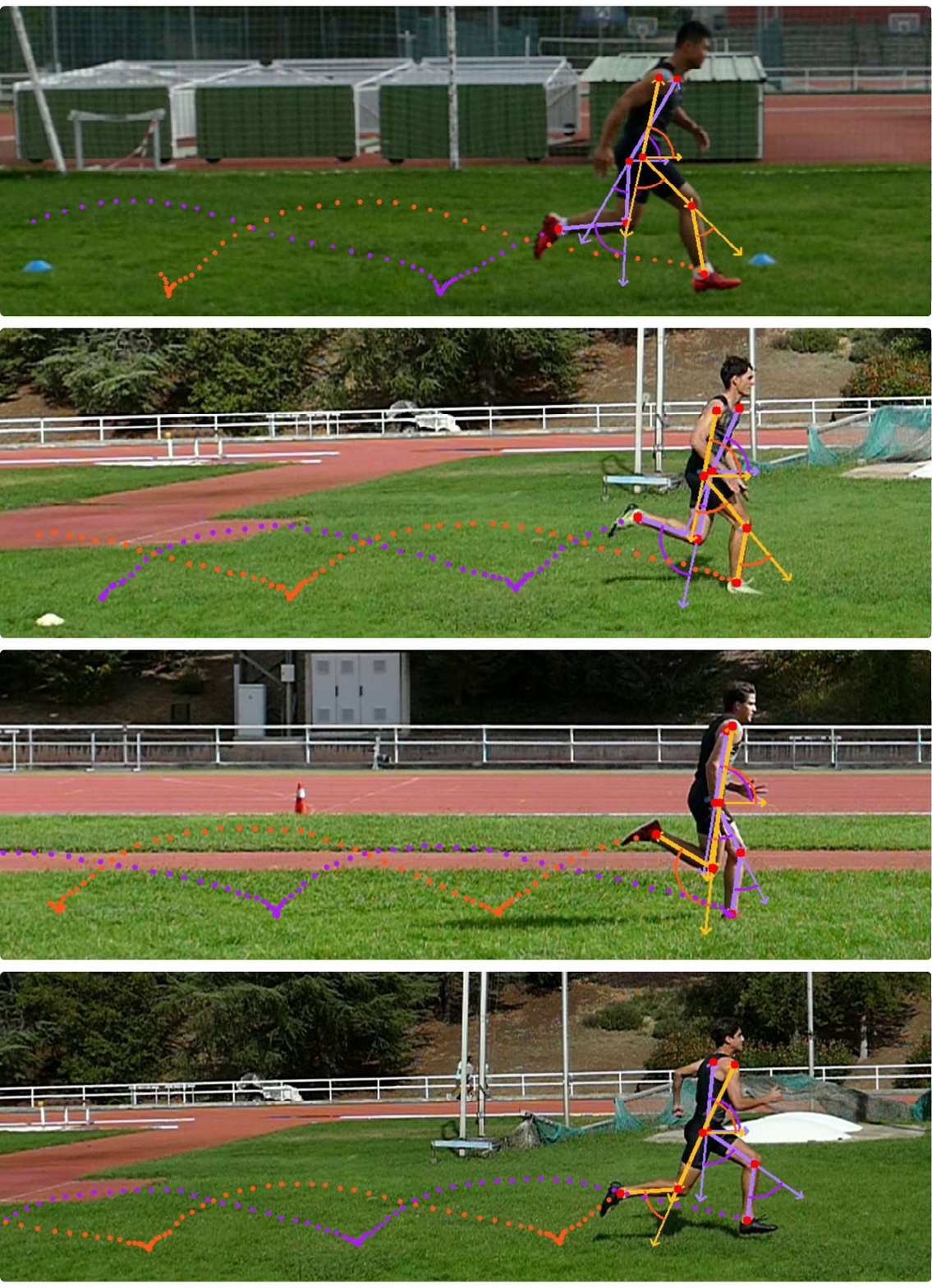}
    \caption{Graphical representation of joint points and angles calculated for four sprinters using the RTMPose model after the proposed tracking post-processing. The orange lines and dots represent the right side of the corridor, while the purple lines and dots represent the left side. (Color image.)}
    \label{fig:JointAnglesIcon}
\end{figure*}

\subsubsection{Biomechanical features generation}

The estimated angles included the trunk's inclination relative to the horizontal plane, the hip angle as the angle between the trunk and thigh, and the knee angle as the angle between the thigh and shank.
For each sprint, the strides are time normalized, starting at 0\% (initial foot strike) and finishing at 100\% (next foot strike of the same foot). Figure~\ref{fig:JointAnglesIcon} shows the evolution of the knee angle over time for four sprinters.
The experimental protocol is oriented towards hip and knee flexion extensions, since the wrong prediction of knee and ankle joint points influences these angles alongside the video camera angle and the occlusions present in the database.

\subsubsection{Performance analysis}

Performance analysis will be conducted using the six trackers and their similarity to ground truth. For this purpose, the root mean squared error will be used as the comparative metric in all experiments. Additionally, the effectiveness of the post-processing methods will be evaluated qualitatively, showing visually their evolution and similarity to the ground truth.

\section{Experiments and Results}


\subsection{ Automatic tracking }

In this first set of experiments, we tested all the trackers considered without post-processing in the VideoRun2D database \cite{garrido2024videorun2d}. 
Table~\ref{tab:beforeMonoResults} shows that trackers based on point-tracking models, such as CoTracker2 and CoTracker3, display notable limitations in the context of sprint tracking, with an obtained root mean squared error (RMSE) of 73.78$^{\circ}$ and 24.97$^{\circ}$ in the six angles.
The results also show that joint trackers exhibit an RMSE lower than 20$^{\circ}$ in almost all cases. The PoseNet tracker is an exception, with errors averaging around 33$^{\circ}$ across the six angles.

\begin{table}[t]
    \caption{Mean error (RMSE) of all the trackers considered (Section~\ref{sec:Tracking}).}
    \resizebox{\columnwidth}{!}{%
    \begin{tabular}{lccccccc}
        \toprule
        \multirow{2}{*}{Tracker}  & \multicolumn{2}{c}{Trunk angle~[°]} & \multicolumn{2}{c}{Hip angle~[°]} & \multicolumn{2}{c}{Knee angle~[°]} & \multirow{2}{*}{Mean[°]} \\
                                & Right & Left & Right & Left & Right & Left        \\
        \midrule
        CoTracker2~\cite{karaev2023cotracker}       & 37.14 & 56.55 & 73.65 & 80.02 & 88.26 & 107.08    &   73.78  \\
        CoTracker3~\cite{karaev2024cotracker3}      & 11.11 & 14.74 & 22.50 & 33.44 & 22.40 & 45.64     &   24.83  \\
        PoseNet~\cite{papandreou2018personlab}      & 8.96  & 10.83 & 32.11 & 30.36 & 47.19 & 73.66     &   33.44  \\
        MediaPipe~\cite{ohri2021device}             & 4.37  & \textbf{4.83}  & 11.41 & 9.46  & 9.93  & 10.57     &   6.33   \\
        MoveNet~\cite{movenet2021}                  & 4.62  & 5.07  & 10.74 & 10.10 & 9.11  & 10.85     &   6.14   \\
        RTMPose~\cite{sun2019deep}                  & \textbf{3.88}  & 5.15  & \textbf{6.67}  & \textbf{8.00}  & \textbf{4.37}  & \textbf{6.15}      &   \textbf{5.62}  \\
        \bottomrule
        \multicolumn{8}{l}{\textbf{Note:} Bold values indicate the best performing.}
    \end{tabular}%
    }
    \label{tab:beforeMonoResults}
\end{table}

Leaving aside the CoTracker2, CoTracker3, and PoseNet trackers, we can observe that the calculation of the trunk angle presents RMSE lower than 12°, but in the hip and knee angles, it presents even more significant errors due to the possible errors obtained in the prediction of the joint points of sprinters (see Figure~\ref{fig:PP_UserStride}). 
Correction of these errors is tested in the following experimental section.

\subsection{ Post-processing tracking }

We now test the tracking post-processing module introduced in Section~\ref{sec:PP_TrackingModule} to correct the different errors in the joint points prediction.
Figure~\ref{fig:PP_UserStride} presents a visual example of the alteration of knee angle over time for the MediaPipe tracker, attributable to a wrong prediction of the joint points. It also presents the correction to the knee angle after implementing the post-processing module.

Moreover, Table~\ref{tab:afterMonoResults} presents the results obtained before and after in each angle, demonstrating improvements in the three trackers.
The MediaPipe and MoveNet trackers, including post-processing, have yielded improved hip and knee angles, as evidenced by a decrease in RMSE from 10.43° to 8.27° for hip angles and from 10.12° to 5.72° for knee angles.
The RTMPose tracker shows a minimal average decrease in RMSE of 0.12°. Finally, the results show that the post-processing module leads to the best 3 trackers (i.e., MediaPipe, MoveNet, and RTMPose) to obtain similar RMSE values, with values approximating 8° for the right and left hip angles, and 5° and 6° for the right and left knee angles, respectively. 
These results demonstrate the utility of the proposed post-processing and the precision achievable by the trackers considered compared to the ground truth.

\begin{table}[tbp]
    \centering
    \caption{RMSE values obtained before and after applying the post-processing module to the joint angles.}
    \resizebox{\columnwidth}{!}{%
    \begin{tabular}{l|cc|cc|cc|cc}
        \toprule
        & \multicolumn{4}{c|}{Hip angle~[°]} & \multicolumn{4}{c}{Knee angle~[°]} \\
        & \multicolumn{2}{c|}{Before} & \multicolumn{2}{c|}{After} & \multicolumn{2}{c|}{Before} & \multicolumn{2}{c}{After} \\
        \cmidrule(lr){2-3} \cmidrule(lr){4-5} \cmidrule(lr){6-7} \cmidrule(lr){8-9}
        Tracker & Right & Left & Right & Left & Right & Left & Right & Left \\
        \midrule
        MediaPipe~\cite{ohri2021device}     & 11.41 & 9.46  & \textbf{8.72} & \textbf{8.01}  & 9.93  & 10.57  & \textbf{5.90} & \textbf{6.16} \\
        MoveNet~\cite{movenet2021}          & 10.74 & 10.10 & \textbf{7.92} & \textbf{8.42}  & 9.11  & 10.85  & \textbf{4.77} & \textbf{6.05} \\
        RTMPose~\cite{sun2019deep}          & 6.67  & 8.00  & \textbf{6.66} & \textbf{8.00}  & 4.37  & 6.15   & \textbf{4.27} & \textbf{5.77} \\
        \bottomrule
        \multicolumn{9}{l}{\textbf{Note:} Bold values indicate the best performing.}
    \end{tabular}
     }
    \label{tab:afterMonoResults}
\end{table}

\begin{figure*}[t!]
    \centering
    \includegraphics[width=0.95\textwidth]{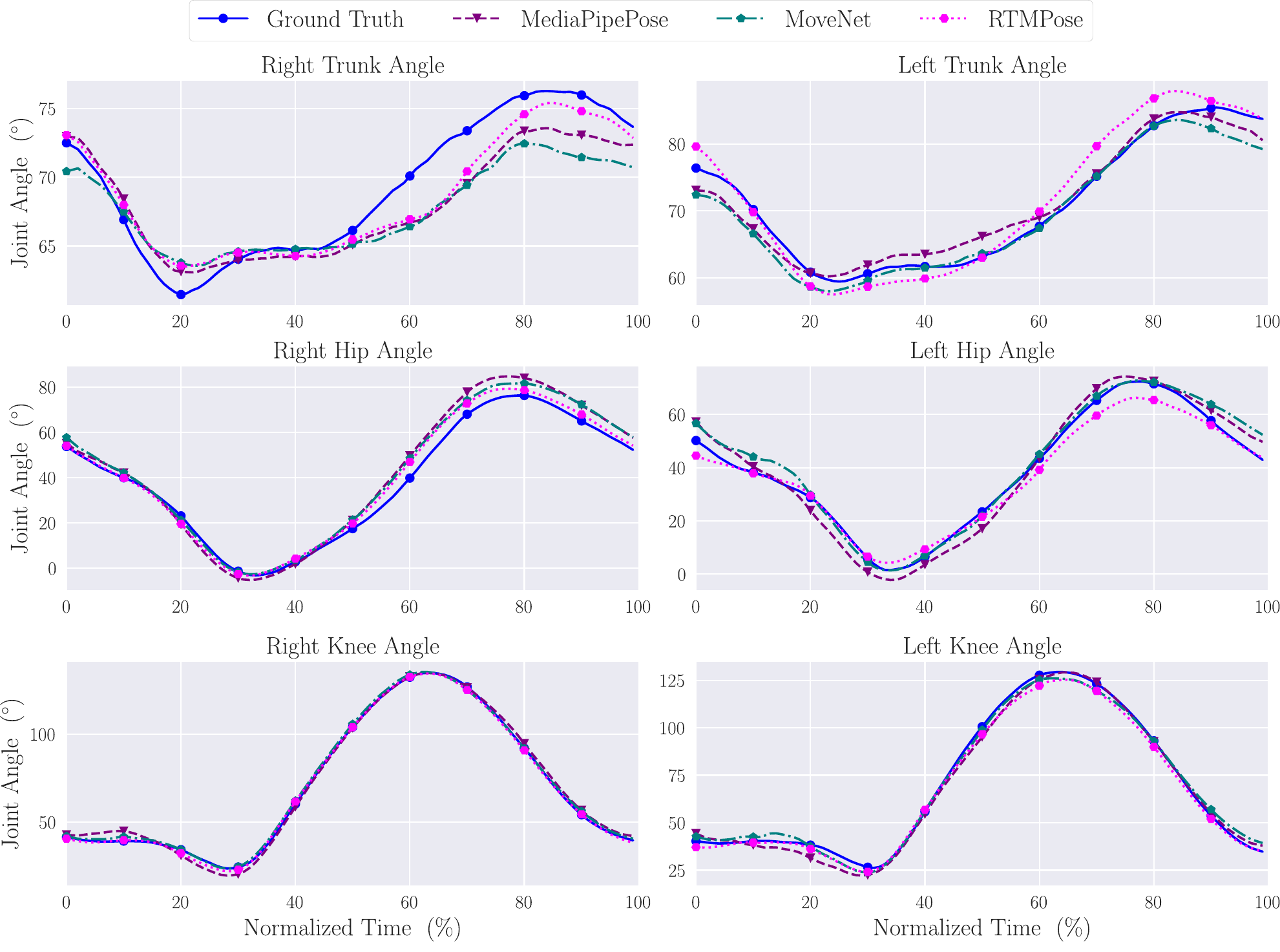}
    \caption{Flex extension pattern of the three joint angles from the best trackers and the ground truth.}
    \label{fig:ExtensionAngles}
\end{figure*}

Figure~\ref{fig:ExtensionAngles} shows the best 3 trackers and their behavior similar to the ground truth in the six angles considered. 
We can see a consistent similarity between the curves for all angles. 
As a result, we observe that the best 3 trackers after post-processing represent very well the ground truth.

\subsection{ Joint angles fusion }

We now experiment a late-fusion strategy, where joint point predictions from the best 3 trackers are averaged to generate a final joint position.

Table~\ref{tab:bestFusion} shows the combination of trackers and demonstrates a substantial reduction in RMSE for the angles of the left trunk, hip, and knee. The combination of MediaPipe, MoveNet, and RTMPose provides the best performance to predict the occluded joint points on the left side.
For the left trunk and knee angles, slight improvements are observed, with an average RMSE reduction of 0.30° in the best case. In contrast, the angle of the left hip shows a more significant improvement of 1.01° in the RMSE (from 8.00° to 6.99°).

These improvements on the left side of the runner suggest that incorporating multiple predictors helps mitigate the occlusion effects caused by the video camera angle during recording.
Furthermore, the RTMPose model demonstrates a notable capacity to predict visible joint points, resulting in satisfactory performance in various joint angle calculations and yielding the lowest RMSE, even without the late fusion.

\begin{table}[t]
    \centering
    \caption{RMSE values obtained using single trackers and the late fusion of trackers.}
    \resizebox{\columnwidth}{!}{%
    \begin{tabular}{l|cc|cc|cc}
        \toprule
            & \multicolumn{2}{c|}{Trunk angle~[°]} & \multicolumn{2}{c|}{Hip angle~[°]} & \multicolumn{2}{c}{Knee angle [°]}   \\
        Tracker/s                          & Right & Left & Right & Left  & Right & Left \\
        \midrule
        MediaPipe+MoveNet+RTMPose   & 4.14            & \textbf{4.51}   & 7.48          & 7.12          & 4.52      & \textbf{5.40}  \\
        MediaPipe+MoveNet          & 4.38            & 4.70            & 8.10          & 7.72          & 4.92      & 5.58          \\
        MediaPipe+RTMPose            & 4.04            & 4.52            & 7.47          & \textbf{6.99} & 4.68      & 5.52          \\
        MoveNet+RTMPose              & 4.12            & 4.70            & 7.09          & 7.45          & 4.33      & 5.56          \\
        \midrule
        MediaPipe                   & 4.37            & 4.83            & 8.72          & 8.01          & 5.90      & 6.16          \\
        MoveNet                     & 4.62            & 5.07            & 7.92          & 8.42          & 4.77      & 6.05          \\
        RTMPose                       & \textbf{3.88}   & 5.15            & \textbf{6.66} & 8.00          & \textbf{4.27}  & 5.77      \\
        \bottomrule
        \multicolumn{7}{l}{\textbf{Note:} Bold values indicate the best performing.}
    \end{tabular}%
    }
    \label{tab:bestFusion}
\end{table}

\section{Discussion and Conclusion}

This study evaluated six state-of-the-art trackers and compared their results with manual labeling for sprint analysis on the VideoRun2D dataset~\cite{garrido2024videorun2d}, which consists of 40 sprints captured from 5 amateur football players (513 full running strides in total).
The different comparative experiments also evaluated a new very useful post-processing module and late fusion strategies to improve the joint prediction of the trackers on the left side of the sprinters due to the occlusion captured by the video camera. 

Of the 6 analyzed trackers, 2 are point-tracking models: CoTracker2 and CoTracker3. Both show a low performance of around 28° RMSE in this task. This is due to the nature of the problem and the objective function used in training~\cite{cronin2024feasibility}. We observed improvements in RMSE between CoTracker3 and CoTracker2, but still both remain significantly inferior to the other trackers evaluated for this task.

On the other hand, the other four trackers are based on joint tracking or skeleton estimation and reduce the RMSE to around 8° after the proposed post-processing module, obtaining joint angle curves similar to ground truth (see Figure~\ref{fig:ExtensionAngles}).
When we look at the joint angles, we see curves similar to those in other related works that use marker-based trackers, such as~\cite{lin2023validity, ota2021verification, nagahara2017kinematics, yang2024improving}, or Inertial Measure Unit (IMU) systems.  
It is important to note that marker-based or IMU systems imply invasive equipment on runners and require significant time and resources for the preparation and training of the operator for data collection. This is in contrast to the systems proposed in this work, which operate in a non-invasive and cost-effective manner~\cite{garrido2024videorun2d,lin2023validity, yang2024improving}.

Note that the angle of capture of the camera significantly impacted the predictability of the articular points of the shoulders and hips. The proximity and persistent occlusion of the same part of the body of the runner contributed to the difficulty in predicting these points. This behavior was also observed here at the angles of the trunk and hips and is also evidenced in related work~\cite{cronin2024feasibility,ota2021verification}.

Finally, the best tracking modules together with the proposed post-processing are competitive compared to more expensive marker-based solutions but still requires improvements. Future work will focus on extending prediction models and combining state-of-the-art trackers~\cite{mmpose2020, fusion18}, the application of user-dependent analysis~\cite{ud05-3,ud05-1,ud05-2}, time-adaptive analysis of different zones of the stride cycle~\cite{fusion18,time18}, and the use of multiple views that augment the current lateral view for improved 2D and 3D information \cite{cava20243dface,cava2025face3d}.

{
    \small
    \bibliographystyle{ieee}
    \bibliography{main}
}

\end{document}